\documentclass{article}
\usepackage[final]{bdl_2018}
\usepackage[utf8]{inputenc} 
\usepackage[T1]{fontenc}    
\usepackage{hyperref}       
\usepackage{url}            
\usepackage{booktabs}       
\usepackage{amsfonts}       
\usepackage{nicefrac}       
\usepackage{microtype}      
\usepackage{mathtools}
\usepackage{graphicx}
\usepackage{wrapfig}
\usepackage{subfigure}
\usepackage{verbatim}

\title{Building robust classifiers through generation of confident out of distribution examples}
\author{
  Kumar Sricharan, Ashok Srivastava \\
  Central Data Science Organization, 
  Intuit Inc. \\
  \texttt{\{sricharan{\textunderscore}kumar,  ashok{\textunderscore}srivastava\}@intuit.com}}

\begin{document}

\maketitle

\begin{abstract}
  Deep learning models are known to be overconfident in their predictions on out of distribution inputs. There have been several pieces of work to address this issue, including a number of approaches for building Bayesian neural networks, as well as closely related work on detection of out of distribution samples. Recently, there has been work on building classifiers that are robust to out of distribution samples by adding a regularization term that maximizes the entropy of the classifier output on out of distribution data. To approximate out of distribution samples (which are not known apriori), a GAN was used for generation of samples at the edges of the training distribution. In this paper, we introduce an alternative GAN based approach for building a robust classifier, where the idea is to use the GAN to explicitly generate out of distribution samples that the classifier is confident on (low entropy), and have the classifier maximize the entropy for these samples. We showcase the effectiveness of our approach relative to state-of-the-art on hand-written characters as well as on a variety of natural image datasets.
\end{abstract}
\section{Introduction}
Deep learning approaches have been shown to be susceptible to being overly confident on unseen instances that are outside of the training distribution. To address this problem, there have been a number of recently proposed techniques. This body of work can be divided into two broad categories: (i) methods that analyze the output of networks trained in a standard fashion to detect out of sample distributions, and (ii) methods that build robust neural networks through modified loss functions including work on Bayesian networks as well as regularized networks. We note that techniques from (i) and (ii) are independent of each other in that techniques from (i) can be combined with techniques from (ii) to produce more effective results. 

The work on using the maximum probability of the classifier~\cite{hendrycks2016baseline}, its modification to use temperature based scaling~\cite{liang2017principled}, and the work on using the underlying feature representations in combination with the Mahalanobis distance~\cite{lee2018simple} belong to category (i). Methods under category (ii) include Bayesian neural networks based on placing prior over weights of the network~\cite{hernandez2015probabilistic}, using dropout as a Bayesian approximation~\cite{li2017dropout}, ensemble based approaches~\cite{lakshminarayanan2017simple} and adding a regularization term to explicitly maximize entropy of out of distribution samples~\cite{lee2017training}. 

Our focus in this work is on the later category. In particular, we build on the recent, state-of-the-art work by Lee \emph{et.al.}~\cite{lee2017training} on generating robust classifiers that don't overfit on out of distribution samples  by explicitly maximizing the uncertainty of a classifier over out of distribution samples. This work in \cite{lee2017training} uses a GAN to generate out-of-distribution samples that are close to training distribution but also simultaneously have high entropy in terms of classifier output over these samples. Their key intuition is that by maximizing the uncertainty of the classifier over these out of distribution samples close to the training distribution, the same effect will be propagated to all samples outside of the training distribution.

\textbf{Contribution} In this paper, we use the same idea of maximizing the classifier uncertainty over out of distribution samples, but take an alternative approach to generating the out of sample distribution. In particular, we use a GAN to generate samples which are \emph{not} from the training distribution that the classifier \emph{is} confident about (i.e., has low entropy), and then have the classifier maximize the uncertainty of the predicted labels over these samples in an iterative fashion. This is in direct contrast to \cite{lee2017training}, where the GAN is used to generate samples close to the training distribution that the classifier is not confident about. This difference can be seen clearly by studying the equations (\ref{eq:gan_orig}) and (\ref{eq:gan_prop}) below. 

This proposed approach leads to improved robust classifiers for out of sample detection because the GAN is explicitly optimized to generate samples away from the training data that the classifier is confident about. These samples can span a much larger space beyond the edges of the training distribution, unlike in \cite{lee2017training}, where the classifier is restricted to maximizing the uncertainty of samples close to the edge of the training distribution only. Henceforth, we will refer to the method in \cite{lee2017training} as BoundaryGAN, and our proposed method as ConfGAN.


\section{Training robust classifiers using GANs}
We start with the confidence loss proposed by Lee \emph{et.al.}\cite{lee2017training}:
    $L_c(\theta) = \mathbb{E}_{P_{in}(\hat{x},\hat{y})} [-\log \mathbb{P}_\theta (y = \hat{y} \mid \hat{x})] + \beta \mathbb{E}_{{P}_{out}(x)}[KL(\mathbb{U}(y) \mid\mid \mathbb{P}_\theta (y \mid {x}))]$
where the first term is the standard cross-entropy loss, and the second term forces the network (parameterized by $\theta$) to generate close to uniform distributions for out of distribution samples (represented by ${P}_{out}$). We note that the proposed loss function corresponds to a Bayesian classifier under a prior on the weights that corresponds to the classifier producing a uniform distribution on the network output of class probabilities $P_\theta(y \mid x)$.

\subsection{GAN for generating boundary samples}
The confidence loss function is very intuitive, but the key difficulty is in determining ${P}_{out}$. In \cite{lee2017training}, they take the approach of approximating ${P}_{out}$ by learning to generate 'boundary' samples at the edges of the training distribution ${P}_{in}$ using a GAN~\cite{goodfellow2014generative}, with the intuition that if the network learns to generate labels distributions with high entropy for these boundary samples, then this will propagate to all out of distribution samples. To generate the boundary out of distribution samples, the generator in BoundaryGAN is asked to learn samples $\bar{x}$ close to the training distribution (via standard GAN loss) which also have low entropy wrt classifier output (via KL divergence loss wrt uniform distribution). In particular, the overall loss function is given by:
\begin{eqnarray}
    \label{eq:gan_orig}
    \min_{\theta} \min_{G} \max_{D} && \underbrace{\mathbb{E}_{P_{in}(\hat{x},\hat{y})} [-\log \mathbb{P}_\theta (y = \hat{y} \mid \hat{x})]}_\text{A} + \underbrace{\beta \mathbb{E}_{{P}_{prior}(z)}[KL(\mathbb{P}_\theta (y \mid G({z}))\mid\mid\mathbb{U}(y))]}_\text{B} \nonumber \\
    && \underbrace{\mathbb{E}_{P_{in}(\hat{x})} [\log D(\hat{x})] + \mathbb{E}_{{P}_{prior}(z)}[\log(1 - D(G(z))]}_\text{C} 
\end{eqnarray}
Here, (A) is the standard cross-entropy loss, and (C) is the standard GAN loss. The term B is introduced in \cite{lee2017training} to both force the generator to generate samples at the boundary of the training distribution, and to regularize the classifier via these out of distribution samples.

\subsection{Proposed method: GAN for generating confident samples}
In contrast to \cite{lee2017training}, we suggest an alternative approach where instead of forcing the generator to learn a distribution where the classifier output is close to uniform, we instead ask the generator to identify samples the classifier is confident about (i.e., low entropy) that are away from the training distribution $P_{in}$. We simultaneously force the classifier to update its parameters $\theta$ to produce probability outputs with low entropy for these samples from the GAN. The overall loss function for our proposed ConfGAN method is given by:
\begin{eqnarray}
    \label{eq:gan_prop}
    \min_{\theta} \underline{\max_{G}} \max_{D} && \underbrace{\mathbb{E}_{P_{in}(\hat{x},\hat{y})} [-\log \mathbb{P}_\theta (y = \hat{y} \mid \hat{x})]}_\text{a} + \underbrace{\beta \mathbb{E}_{{P}_{prior}(z)}[KL(\mathbb{P}_\theta (y \mid G({z}))\mid\mid\mathbb{U}(y))]}_\text{b} \nonumber \\
    && \underbrace{\mathbb{E}_{P_{in}(\hat{x})} [\log D(\hat{x})] + \mathbb{E}_{{P}_{prior}(z)}[\log(1 - D(G(z))]}_\text{c} 
\end{eqnarray}
This is similar to the loss function (Eq~\ref{eq:gan_orig}) for BoundaryGAN with the only difference being that in BoundaryGAN, the overall loss is minimized with respect to the generator $G$ instead of being maximized as in Eq.\ref{eq:gan_prop}. For this reason, in term (b), we use the reverse KL divergence instead of the forward version used in~\cite{lee2017training}. In practice, for term (b), we use the reverse KL term for maximization wrt the generator, and the original forward KL term from ~\cite{lee2017training} for minimization wrt the classifier. This is because the forward KL terms supplies stronger gradients that push the predicted distribution $\mathbb{P}_\theta (y = \hat{y} \mid \hat{x})$ towards the uniform distribution.

Asking the generator to maximize the loss function has the effect of forcing the generator to generate samples the classifier is confident about through term (b), and at the same time, are not from the training distribution through term (c). 

\section{Experimental results}
While the modification from the original loss in BoundaryGAN to the proposed loss function is easy to implement, the effect of the modification is significant in terms of the results. We show this through experimental results on two different data sets - handwritten characters, and natural images. For measuring out-of-distribution detection accuracy, as in ~\cite{lee2017training}, we use threshold-based detectors~\cite{hendrycks2016baseline} that computes the maximum value of predictive distribution on a test sample and classifies it as  in-distribution if this value is above some threshold. 

\subsection{Handwritten characters}
\begin{wrapfigure}{r}{0.53\textwidth}
\vspace{-0.25in}
\begin{minipage}[b][0.18\textheight][s]{0.18\textwidth}
  \centering
  \includegraphics[height=0.07\textheight,width=\textwidth]{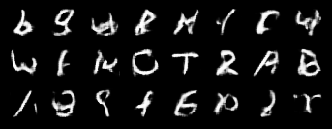}
  {\small{ (a) BoundaryGAN}}
  \vfill
  \includegraphics[height=0.07\textheight,width=\textwidth]{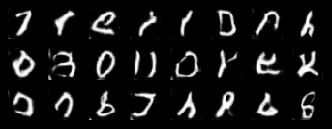}
  {\small{(b) ConfGAN}}
\end{minipage}
\includegraphics[height=0.18\textheight,width=.346\textwidth]{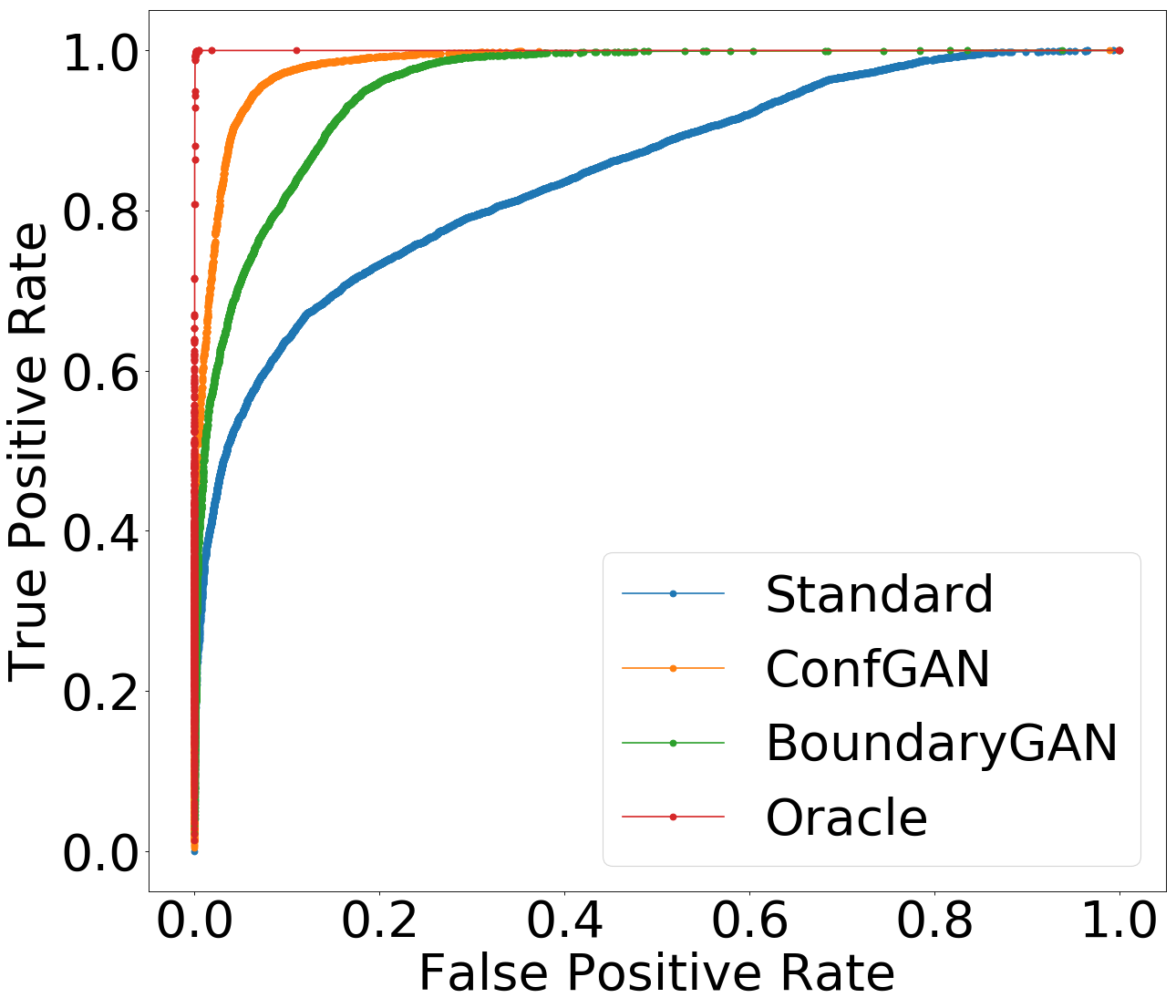}
\end{wrapfigure}
In the first experiment, we use the handwritten letters from EMNIST~\cite{cohen2017emnist} as the in-sample distribution, and the FashionMNIST dataset~\cite{xiao2017fashion} as the out of sample distribution. The neural network for the classifier is LeNet-5~\cite{lecun1995comparison} modified to support 47 classes, and the discriminator and generator use the DCGAN architecture~\cite{radford2015unsupervised}. We use $\beta=1$. We showcase two sets of results: first, we compare the samples of the generator from BoundaryGAN with ConfGAN. As can be seen from the figure, BoundaryGAN produces samples that look like complete characters similar to the actual EMNIST data (similar to Figure 3(d) in \cite{lee2017training}), whereas ConfGAN produces images that look like well-defined, but incomplete parts of characters - i.e., samples that are not from the training distribution that a classifier is likely to be overconfident about. Next, we compare the ROC curves measuring detection performance of BoundaryGAN, ConfGAN, the standard classifier trained with no regularization (lower bound), and an oracle classifier that is trained with FashionMNIST as the out of distribution data (upper bound). From Figure~\ref{fig:nat_img_comp}, we can see that ConfGAN comes close to the performance of the oracle, and significantly outperforms BoundaryGAN.

\subsection{Image datasets using VGG}
\begin{figure}[!h]
  \includegraphics[width=\linewidth]{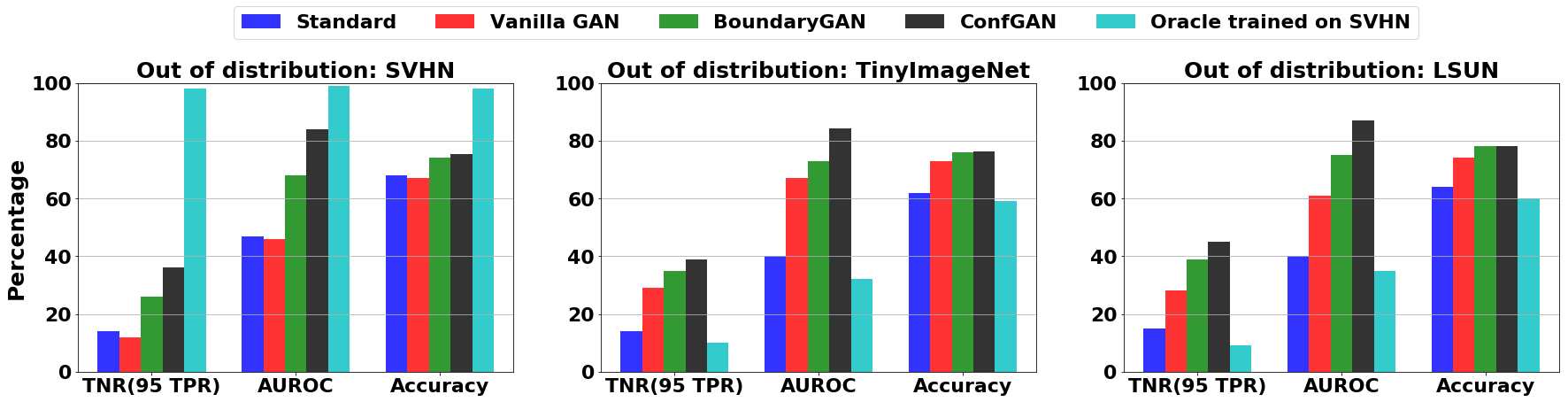}
  \label{fig:nat_img_comp}
\end{figure}

The second experiment is on a combination of various datasets: CIFAR~\cite{krizhevsky2009learning}, SVHN~\cite{netzer2011reading}, ImageNet~\cite{deng2009imagenet}, and LSUN~\cite{yu2015lsun}. We train VGGNet~\cite{szegedy2015going} for classifying CIFAR-10, use DCGAN for the generator and discriminator, with $\beta=0.1$, and use datasets from \{SVHN, ImageNet, LSUN\} as the out of sample datasets. Under this setup, we measure out of sample detection performance using the same evaluation metrics in ~\cite{lee2017training} (TNR at 95\% TPR, AUROC, Detection accuracy). From Figure~\ref{fig:nat_img_comp}, we see that the proposed ConfGAN uniformly outperforms BoundaryGAN and other baselines (except oracle trained and applied on SVHN). The performance improvement of ConfGAN over BoundaryGAN is especially apparent wrt AUROC (with an average improvement of $\sim$ 12\%) across the three datasets.

\section{Conclusion}
We proposed a new method for building robust classifiers to detect out of distribution samples. The method is based on using a GAN to generate samples away from the training data that the classifier is confident about, and have the classifier maximize the uncertainty over these points. Our method outperforms several baselines in this space on handwritten character and natural image datasets.

\bibliographystyle{plain}
\bibliography{ref}

\begin{thebibliography}{10}

\bibitem{cohen2017emnist}
Gregory Cohen, Saeed Afshar, Jonathan Tapson, and Andr{\'e} van Schaik.
\newblock Emnist: an extension of mnist to handwritten letters.
\newblock {\em arXiv preprint arXiv:1702.05373}, 2017.

\bibitem{deng2009imagenet}
Jia Deng, Wei Dong, Richard Socher, Li-Jia Li, Kai Li, and Li~Fei-Fei.
\newblock Imagenet: A large-scale hierarchical image database.
\newblock In {\em Computer Vision and Pattern Recognition, 2009. CVPR 2009.
  IEEE Conference on}, pages 248--255. Ieee, 2009.

\bibitem{goodfellow2014generative}
Ian Goodfellow, Jean Pouget-Abadie, Mehdi Mirza, Bing Xu, David Warde-Farley,
  Sherjil Ozair, Aaron Courville, and Yoshua Bengio.
\newblock Generative adversarial nets.
\newblock In {\em Advances in neural information processing systems}, pages
  2672--2680, 2014.

\bibitem{hendrycks2016baseline}
Dan Hendrycks and Kevin Gimpel.
\newblock A baseline for detecting misclassified and out-of-distribution
  examples in neural networks.
\newblock {\em arXiv preprint arXiv:1610.02136}, 2016.

\bibitem{hernandez2015probabilistic}
Jos{\'e}~Miguel Hern{\'a}ndez-Lobato and Ryan Adams.
\newblock Probabilistic backpropagation for scalable learning of bayesian
  neural networks.
\newblock In {\em International Conference on Machine Learning}, pages
  1861--1869, 2015.

\bibitem{krizhevsky2009learning}
Alex Krizhevsky and Geoffrey Hinton.
\newblock Learning multiple layers of features from tiny images.
\newblock Technical report, Citeseer, 2009.

\bibitem{lakshminarayanan2017simple}
Balaji Lakshminarayanan, Alexander Pritzel, and Charles Blundell.
\newblock Simple and scalable predictive uncertainty estimation using deep
  ensembles.
\newblock In {\em Advances in Neural Information Processing Systems}, pages
  6402--6413, 2017.

\bibitem{lecun1995comparison}
Yann LeCun, LD~Jackel, Leon Bottou, A~Brunot, Corinna Cortes, JS~Denker, Harris
  Drucker, I~Guyon, UA~Muller, Eduard Sackinger, et~al.
\newblock Comparison of learning algorithms for handwritten digit recognition.
\newblock In {\em International conference on artificial neural networks},
  volume~60, pages 53--60. Perth, Australia, 1995.

\bibitem{lee2017training}
Kimin Lee, Honglak Lee, Kibok Lee, and Jinwoo Shin.
\newblock Training confidence-calibrated classifiers for detecting
  out-of-distribution samples.
\newblock {\em arXiv preprint arXiv:1711.09325}, 2017.

\bibitem{lee2018simple}
Kimin Lee, Kibok Lee, Honglak Lee, and Jinwoo Shin.
\newblock A simple unified framework for detecting out-of-distribution samples
  and adversarial attacks.
\newblock {\em arXiv preprint arXiv:1807.03888}, 2018.

\bibitem{li2017dropout}
Yingzhen Li and Yarin Gal.
\newblock Dropout inference in bayesian neural networks with alpha-divergences.
\newblock {\em arXiv preprint arXiv:1703.02914}, 2017.

\bibitem{liang2017principled}
Shiyu Liang, Yixuan Li, and R~Srikant.
\newblock Principled detection of out-of-distribution examples in neural
  networks. arxiv preprint.
\newblock {\em arXiv preprint arXiv:1706.02690}, 2017.

\bibitem{netzer2011reading}
Yuval Netzer, Tao Wang, Adam Coates, Alessandro Bissacco, Bo~Wu, and Andrew~Y
  Ng.
\newblock Reading digits in natural images with unsupervised feature learning.
\newblock In {\em NIPS workshop on deep learning and unsupervised feature
  learning}, volume 2011, page~5, 2011.

\bibitem{radford2015unsupervised}
Alec Radford, Luke Metz, and Soumith Chintala.
\newblock Unsupervised representation learning with deep convolutional
  generative adversarial networks.
\newblock {\em arXiv preprint arXiv:1511.06434}, 2015.

\bibitem{szegedy2015going}
Christian Szegedy, Wei Liu, Yangqing Jia, Pierre Sermanet, Scott Reed, Dragomir
  Anguelov, Dumitru Erhan, Vincent Vanhoucke, and Andrew Rabinovich.
\newblock Going deeper with convolutions.
\newblock In {\em Proceedings of the IEEE conference on computer vision and
  pattern recognition}, pages 1--9, 2015.

\bibitem{xiao2017fashion}
Han Xiao, Kashif Rasul, and Roland Vollgraf.
\newblock Fashion-mnist: a novel image dataset for benchmarking machine
  learning algorithms.
\newblock {\em arXiv preprint arXiv:1708.07747}, 2017.

\bibitem{yu2015lsun}
Fisher Yu, Ari Seff, Yinda Zhang, Shuran Song, Thomas Funkhouser, and Jianxiong
  Xiao.
\newblock Lsun: Construction of a large-scale image dataset using deep learning
  with humans in the loop.
\newblock {\em arXiv preprint arXiv:1506.03365}, 2015.

\end{thebibliography}

\end{document}